\documentclass[11pt,a4paper]{article}
\usepackage[hyperref]{acl2018}
\usepackage{times}
\usepackage{latexsym}
\usepackage{graphicx}
\usepackage{booktabs}
\usepackage{multirow}
\usepackage{CJKutf8}
\usepackage[flushleft]{threeparttable}
\usepackage{amsmath}

\usepackage{url}

\aclfinalcopy 

\title{Bilingual Character Representation for Efficiently Addressing Out-of-Vocabulary Words in Code-Switching Named Entity Recognition}

\author{Genta Indra Winata, Chien-Sheng Wu, Andrea Madotto, Pascale Fung \\
        Center for Artificial Intelligence Research (CAiRE)\\
        Department of Electronic and Computer Engineering\\
        Hong Kong University of Science and Technology, Clear Water Bay, Hong Kong\\ 
        \tt \{giwinata, cwuak, eeandreamad\}@ust.hk, pascale@ece.ust.hk}
        
\date{}

\begin{document}
\begin{CJK*}{UTF8}{gbsn}

\maketitle
\begin{abstract}
We propose an LSTM-based model with hierarchical architecture on named entity recognition from code-switching Twitter data. Our model uses bilingual character representation and transfer learning to address out-of-vocabulary words. In order to mitigate data noise, we propose to use token replacement and normalization. In the 3rd Workshop on Computational Approaches to Linguistic Code-Switching Shared Task, we achieved second place with 62.76\% harmonic mean F1-score for English-Spanish language pair without using any gazetteer and knowledge-based information.
\end{abstract}

\section{Introduction}
Named Entity Recognition (NER) predicts which word tokens refer to location, people, organization, time, and other entities from a word sequence. Deep neural network models have successfully achieved the state-of-the-art performance in NER tasks \cite{cohenmulti, chiu2016named, lample2016neural, shen2017deep} using monolingual corpus. However, learning from code-switching tweets data is very challenging due to several reasons: (1) words may have different semantics in different context and language, for instance, the word ``cola'' can be associated with product or ``queue'' in Spanish (2) data from social media are noisy, with many inconsistencies such as spelling mistakes, repetitions, and informalities which eventually points to Out-of-Vocabulary (OOV) words issue (3) entities may appear in different language other than the matrix language. For example ``todos los Domingos en Westland Mall" where  ``Westland Mall" is an English named entity. 

Our contributions are two-fold: (1) bilingual character bidirectional RNN is used to capture character-level information and tackle OOV words issue (2) we apply transfer learning from monolingual pre-trained word vectors to adapt the model with different domains in a bilingual setting. In our model, we use LSTM to capture long-range dependencies of the word sequence and character sequence in bilingual character RNN. In our experiments, we show the efficiency of our model in handling OOV words and bilingual word context.

\section{Related Work}
Convolutional Neural Network (CNN) was used in NER task as word decoder by \citet{collobert2011natural} and a few years later, \citet{huang2015bidirectional} introduced Bidirectional Long-Short Term Memory (BiLSTM) \cite{sundermeyer2012lstm}. Character-level features were explored by using neural architecture and replaced hand-crafted features \cite{dyer2015transition, lample2016neural, chiu2016named, limsopatham2016bidirectional}. \citet{lample2016neural} also showed Conditional Random Field (CRF) \cite{lafferty2001conditional} decoders to improve the results and used Stack memory-based LSTMs for their work in sequence chunking. \citet{aguilar2017multi} proposed multi-task learning by combining Part-of-Speech tagging task with NER and using gazetteers to provide language-specific knowledge. Character-level embeddings were used to handle the OOV words problem in NLP tasks such as NER \cite{lample2016neural}, POS tagging, and language modeling \cite{ling2015finding}.

\section{Methodology}

\begin{table*}[!htb]
\centering
\caption{OOV words rates on ENG-SPA dataset before and after preprocessing}
\label{eng-spa-oov-statistics}
\begin{tabular}{@{}lccccc@{}}
\hline
\multicolumn{1}{c}{\multirow{2}{*}{\textbf{}}}                                   & \multicolumn{2}{|c|}{\textbf{Train}}                                  & \multicolumn{2}{c}{\textbf{Dev}}                                             & \multicolumn{1}{|c}{\textbf{Test}}                             \\ \cline{2-6} 
\multicolumn{1}{c}{}                                                             & \multicolumn{1}{|c|}{All}                              & Entity                           & \multicolumn{1}{|c|}{All}                              & Entity                           & \multicolumn{1}{|c}{All}                              \\ \hline
Corpus & \multicolumn{1}{|c|}{-} & - & \multicolumn{1}{|c|}{18.91\%} & 31.84\% & \multicolumn{1}{|c}{49.39\%} \\ \hline
FastText (eng) \cite{mikolov2018advances} & \multicolumn{1}{|c|}{62.62\%} & 16.76\% & \multicolumn{1}{|c|}{19.12\%} & 3.91\% & \multicolumn{1}{|c}{54.59\%} \\ \hline
+ FastText (spa) \cite{grave2018learning} & \multicolumn{1}{|c|}{49.76\%} & 12.38\% & \multicolumn{1}{|c|}{11.98\%} & 3.91\% & \multicolumn{1}{|c}{39.45\%} \\ \hline
+ token replacement & \multicolumn{1}{|c|}{12.43\%} & 12.35\% & \multicolumn{1}{|c|}{7.18\%} & 3.91\% & \multicolumn{1}{|c}{9.60\%}  \\ \hline
\textbf{+ token normalization} & \multicolumn{1}{|c|}{\textbf{7.94\%}} & \textbf{8.38\%} & \multicolumn{1}{|c|}{\textbf{5.01\%}} & \textbf{1.67\%} & \multicolumn{1}{|c}{\textbf{6.08\%}}  \\ \hline
\end{tabular}
\end{table*}

\subsection{Dataset} 
For our experiment, we use English-Spanish (ENG-SPA) Tweets data from Twitter provided by \citet{calcs2018shtask}. There are nine different named-entity labels. The labels use IOB format (Inside, Outside, Beginning) where every token is labeled as \textnormal{\tt B-label} in the beginning and follows with \textnormal{\tt I-label} if it is inside a named entity, or \textnormal{\tt O} otherwise. For example ``Kendrick Lamar'' is represented as \textnormal{\tt B-PER I-PER}. Table \ref{data-statistics-eng-spa} and Table \ref{data-statistics-eng-spa-named-entities} show the statistics of the dataset.

\begin{table}[!htb]
\centering
\caption{Data Statistics for ENG-SPA Tweets}
\label{data-statistics-eng-spa}
\begin{tabular}{@{}llll@{}}
\hline
& \multicolumn{1}{|c|}{\textbf{Train}} & \multicolumn{1}{c}{\textbf{Dev}} & \multicolumn{1}{|c}{\textbf{Test}} \\ \hline
\# Words & \multicolumn{1}{|c|}{616,069} & 9,583 & \multicolumn{1}{|c}{183,011} \\ \hline
\end{tabular}
\end{table}

\begin{table}[!htb]
\centering
\caption{Entity Statistics for ENG-SPA Tweets}
\label{data-statistics-eng-spa-named-entities}
\begin{tabular}{@{}rcc@{}}
\hline
\multicolumn{1}{r}{\textbf{Entities}} & \multicolumn{1}{|c|}{\textbf{Train}} & \multicolumn{1}{c}{\textbf{Dev}}  \\ \hline
\# Person & \multicolumn{1}{|c|}{4701} & 75 \\ \hline
\# Location & \multicolumn{1}{|c|}{2810} & 10 \\ \hline
\# Product & \multicolumn{1}{|c|}{1369} & 16 \\ \hline
\# Title & \multicolumn{1}{|c|}{824} & 22 \\ \hline
\# Organization & \multicolumn{1}{|c|}{811} & 9 \\ \hline
\# Group & \multicolumn{1}{|c|}{718} & 4 \\ \hline
\# Time & \multicolumn{1}{|c|}{577} & 6 \\ \hline
\# Event & \multicolumn{1}{|c|}{232} & 4 \\ \hline
\# Other & \multicolumn{1}{|c|}{324} & 6 \\ \hline
\end{tabular}
\end{table}

``Person'', ``Location'', and ``Product'' are the most frequent entities in the dataset, and the least common ones are ``Time", ``Event", and ``Other'' categories. `Other'' category is the least trivial among all because it is not well clustered like others. 

\subsection{Feature Representation}
In this section, we describe word-level and character-level features used in our model.

\textbf{Word Representation: } Words are encoded into continuous representation. The vocabulary is built from training data. The Twitter data are very noisy, there are many spelling mistakes, irregular ways to use a word and repeating characters. We apply several strategies to overcome the issue. We use 300-dimensional English \cite{mikolov2018advances} and Spanish \cite{grave2018learning} FastText pre-trained word vectors which comprise two million words vocabulary each and they are trained using Common Crawl and Wikipedia. To create the shared vocabulary, we concatenate English and Spanish word vectors.

For preprocessing, we propose the following steps: 
\begin{enumerate}
\item \textbf{Token replacement: } Replace user hashtags (\#user) and mentions (@user) with ``USR", and URL (https://domain.com) with ``URL".
\item \textbf{Token normalization: } Concatenate Spanish and English FastText word vector vocabulary. Normalize OOV words by using one out of these heuristics and check if the word exists in the vocabulary sequentially
	\begin{enumerate}
    \item Capitalize the first character
    \item Lowercase the word
    \item Step (b) and remove repeating characters, such as \textit{``hellooooo"} into \textit{``hello"} or \textit{``lolololol"} into \textit{``lol"}
    \item Step (a) and (c) altogether
    \end{enumerate}
\end{enumerate}


Then, the effectiveness of the preprocessing and transfer learning in handling OOV words are analyzed. The statistics is showed in Table \ref{eng-spa-oov-statistics}. It is clear that using FastText word vectors reduce the OOV words rate especially when we concatenate the vocabulary of both languages. Furthermore, the preprocessing strategies dramatically decrease the number of unknown words.

\textbf{Character Representation: } We concatenate all possible characters for English and Spanish, including numbers and special characters. English and Spanish have most of the characters in common, but, with some additional unique Spanish characters. All cases are kept as they are.

\subsection{Model Description}
In this section, we describe our model architecture and hyper-parameters setting.

\begin{figure}[!htb]
  \centering
  \includegraphics[width=1.0\linewidth]{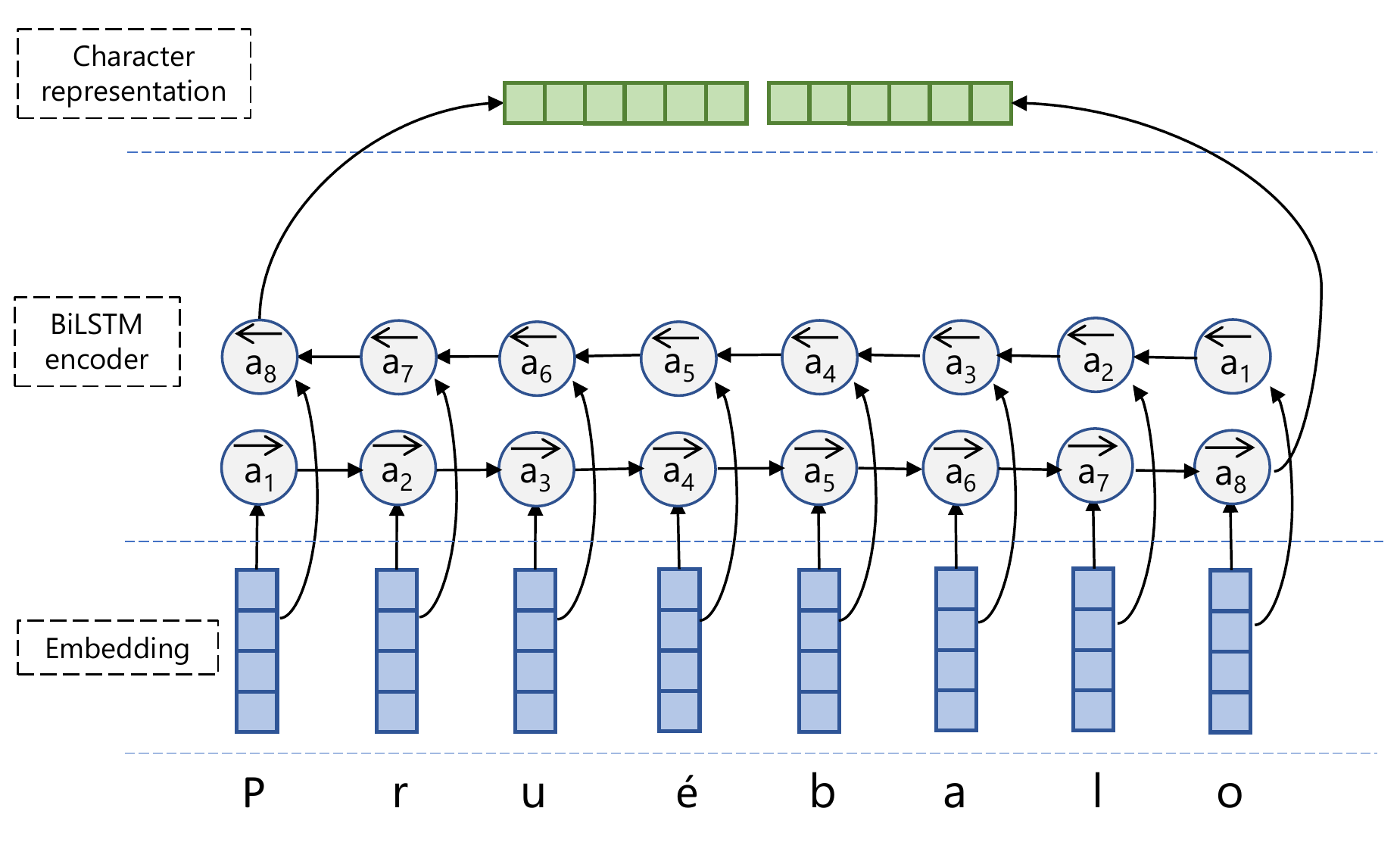}
  \caption{Bilingual Char-RNN architecture}
  \label{fig:char-rnn}
\end{figure}

\textbf{Bilingual Char-RNN: } This is one of the approaches to learn character-level embeddings without needing of any lexical hand-crafted features. We use an RNN for representing the word with character-level information \cite{lample2016neural}. Figure \ref{fig:char-rnn} shows the model architecture. The inputs are characters extracted from a word and every character is embedded with $d$ dimension vector. Then, we use it as the input for a Bidirectional LSTM as character encoder, wherein every time step, a character is input to the network. Consider $a_t$ as the hidden states for word $t$.
\[ a_t = (a_1^1, a_t^2, ..., a_t^\textnormal{V}) \]

where $\textnormal{V}$ is the character length. The representation of the word is obtained by taking $a_t^\textnormal{V}$ which is the last hidden state. 

\begin{figure}[!htb]
  \centering
  \includegraphics[width=1.0\linewidth]{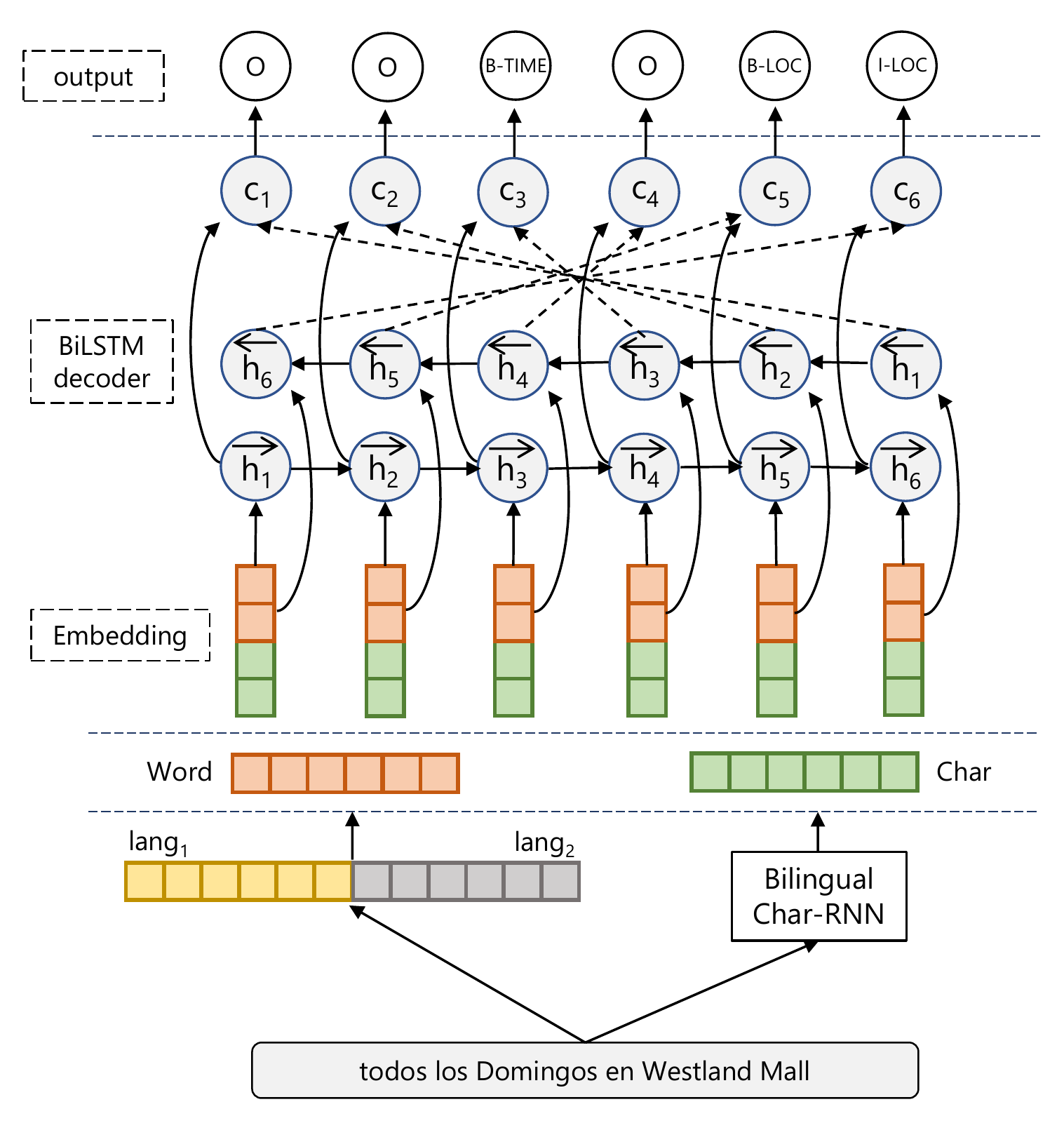}
  \caption{Main architecture}
  \label{fig:overall}
\end{figure}

\textbf{Main Architecture: } Figure \ref{fig:overall} presents the overall architecture of the system. The input layers receive word and character-level representations from English and Spanish pre-trained FastText word vectors and Bilingual Char-RNN. Consider $\textbf{X}$ as the input sequence:
\[ \textbf{X} = {(x_1, x_2, ..., x_\textnormal{N})} \]

where N is the length of the sequence. We fix the word embedding parameters. Then, we concatenate both vectors to get a richer word representation $u_t$. Afterwards, we pass the vectors to bidirectional LSTM. 
\[ u_t = x_t \oplus a_t \]
\[ \overrightarrow{h_t} = \overrightarrow{\textnormal{LSTM}}(u_t, \overrightarrow{h_{t-1}})\textnormal{, } \overleftarrow{h_t} = \overleftarrow{\textnormal{LSTM}}(u_t, \overleftarrow{h_{t-1}})  \]
\[ c_t = \overrightarrow{h_t} \oplus \overleftarrow{h_t} \]

where $\oplus$ denotes the concatenation operator. Dropout is applied to the recurrent layer. At each time step we make a prediction for the entity of the current token. A softmax function is used to calculate the probability distribution of all  possible named-entity tags.
\[ y_t = \frac{e^{c_t}}{\sum_{j=1}^T e^{c_j}} \textnormal{, where } j = 1 \textnormal{, .., T} \]

where $\textnormal{y}_t$ is the probability distribution of tags at word $t$ and $\textnormal{T}$ is the maximum time step. Since there is a variable number of sequence length, we padded the sequence and applied mask when calculating cross-entropy loss function. Our model does not use any gazetteer and knowledge-based information, and it can be easily adapted to another language pair.

\begin{table*}[!htb]
\centering
\caption{Results on ENG-SPA Dataset ($\ddagger$ result(s) from the shared task organizer \cite{calcs2018shtask} \hspace{1mm} $\dagger$ without token normalization) }
\label{results-eng-spa}
\begin{tabular}{@{}llcc@{}}
\hline
\multicolumn{1}{c}{\textbf{Model}}                    & \multicolumn{1}{|c|}{\textbf{Features}} & \textbf{\begin{tabular}[c]{@{}c@{}}F1\\ Dev\end{tabular}} & \multicolumn{1}{|c}{\textbf{\begin{tabular}[c]{@{}c@{}}F1\\ Test\end{tabular}}} \\ \hline
Baseline$^\ddagger$                                    & \multicolumn{1}{|c|}{Word}                                  & -                                                         & \multicolumn{1}{|c}{53.2802\%}                                                  \\ \hline
BiLSTM$^\dagger$                                      & \multicolumn{1}{|c|}{Word + Char-RNN}                           & 46.9643\%                                                 & \multicolumn{1}{|c}{53.4759\%}                                                  \\ \hline
BiLSTM                                                & \multicolumn{1}{|c|}{FastText (eng)}                        & 57.7174\%                                                 & \multicolumn{1}{|c}{59.9098\%}                                                  \\ \hline
BiLSTM                                                & \multicolumn{1}{|c|}{FastText (eng-spa)}                    & 57.4177\%                                                 & \multicolumn{1}{|c}{60.2426\%}                                                  \\ \hline
BiLSTM                                                &  \multicolumn{1}{|c|}{+ Char-RNN}               & 65.2217\%                                                 & \multicolumn{1}{|c}{61.9621\%}                                                  \\ \hline
+ post                                         & \multicolumn{1}{|c|}{}                                      & \textbf{65.3865\%}                                        & \multicolumn{1}{|c}{\textbf{62.7608\%}}                                      \\ \hline \hline
\multicolumn{4}{l}{\textbf{Competitors$^\ddagger$}}                                                                                                                                                                               \\ \hline
IIT BHU ($1^{st}$ place)                              & \multicolumn{1}{|c|}{-}                                     & -                                                         & \multicolumn{1}{|c}{63.7628\% (+1.0020\%)}                                      \\ \hline
FAIR \hspace{4.4mm} ($3^{rd}$ place) & \multicolumn{1}{|c|}{-}                                     & -                                                         & \multicolumn{1}{|c}{62.6671\% (- 0.0937\%)}                                     \\ \hline
\end{tabular}
\end{table*}

\subsection{Post-processing}
\begin{figure}[!htb]
  \centering
  \includegraphics[width=1.0\linewidth]{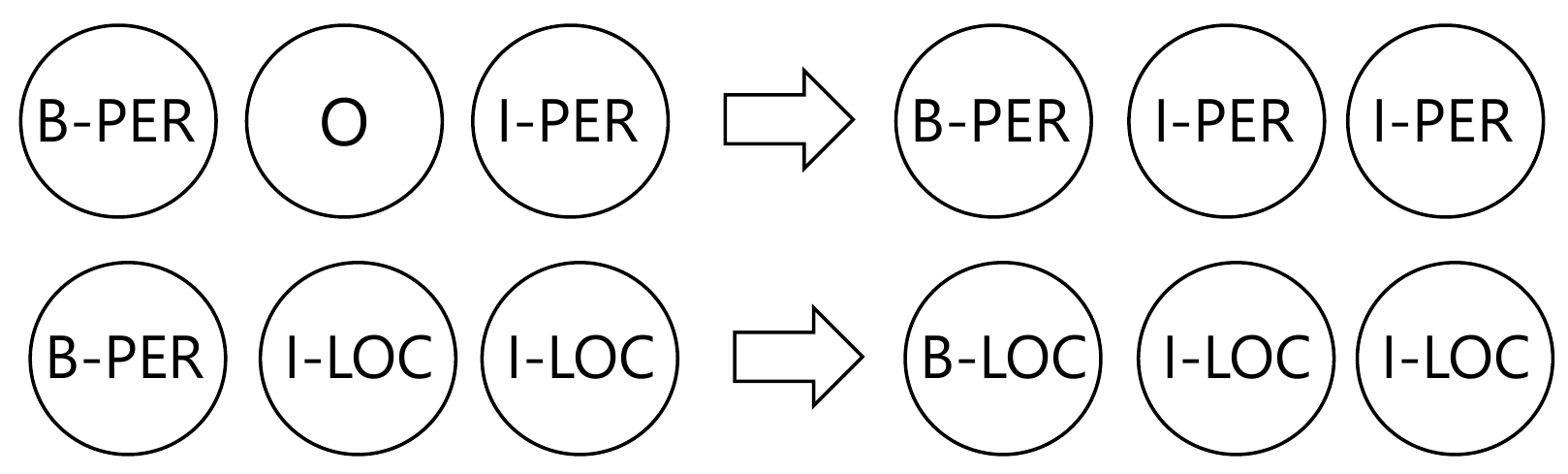}
  \caption{Post-processing examples}
  \label{fig:post1}
\end{figure}

We found an issue during the prediction where some words are labeled with \textnormal{\tt O}, in between \textnormal{\tt B-label} and \textnormal{\tt I-label} tags. Our solution is to insert \textnormal{\tt I-label} tag if the tag is surrounded by \textnormal{\tt B-label} and \textnormal{\tt I-label} tags with the same entity category. Another problem we found that many \textnormal{\tt I-label} tags are paired with \textnormal{\tt B-label} in different categories. So, we replace \textnormal{\tt B-label} category tag with corresponding \textnormal{\tt I-label} category tag. This step improves the result of the prediction on the development set. Figure \ref{fig:post1} shows the examples.

\subsection{Experimental Setup}
We trained our LSTM models with a hidden size of 200. We used batch size equals to 64. The sentences were sorted by length in descending order. Our embedding size is 300 for word and 150 for characters. Dropout \cite{srivastava2014dropout} of 0.4 was applied to all LSTMs. Adam Optimizer was chosen with an initial learning rate of 0.01. We applied time-based decay of $\sqrt{2}$ decay rate and stop after two consecutive epochs without improvement. We tuned our model with the development set and evaluated our best model with the test set using harmonic mean F1-score metric with the script provided by \citet{calcs2018shtask}.

\section{Results}

Table \ref{results-eng-spa} shows the results for ENG-SPA tweets. Adding pre-trained word vectors and character-level features improved the performance. Interestingly, our initial attempts at adding character-level features did not improve the overall performance, until we apply dropout to the Char-RNN. The performance of the model improves significantly after transfer learning with FastText word vectors while it also reduces the number of OOV words in the development and test set. The margin between ours and first place model is small, approximately 1\%.

We try to use sub-words representation from Spanish FastText \cite{grave2018learning}, however, it does not improve the result since the OOV words consist of many special characters, for example, \textit{``/IAtrevido/Provocativo", ``Twets/wek"}, and possibly create noisy vectors and most of them are not entity words. 

\section{Conclusion}
This paper presents a bidirectional LSTM-based model with hierarchical architecture using bilingual character RNN to address the OOV words issue. Moreover, token replacement, token normalization, and transfer learning reduce OOV words rate even further and significantly improves the performance. The model achieved 62.76\% F1-score for English-Spanish language pair without using any gazetteer and knowledge-based information.

\section*{Acknowledgments}
This work is partially funded by ITS/319/16FP of the Innovation Technology Commission, HKUST 16214415 \& 16248016 of Hong Kong Research Grants Council, and RDC 1718050-0 of EMOS.AI.


\bibliography{acl2018}

\begin{thebibliography}{16}
\expandafter\ifx\csname natexlab\endcsname\relax\def\natexlab#1{#1}\fi

\bibitem[{Aguilar et~al.(2018)Aguilar, AlGhamdi, Soto, Diab, Hirschberg, and
  Solorio}]{calcs2018shtask}
Gustavo Aguilar, Fahad AlGhamdi, Victor Soto, Mona Diab, Julia Hirschberg, and
  Thamar Solorio. 2018.
\newblock {Overview of the CALCS 2018 Shared Task: Named Entity Recognition on
  Code-switched Data}.
\newblock In \emph{Proceedings of the Third Workshop on Computational
  Approaches to Linguistic Code-Switching}, Melbourne, Australia. Association
  for Computational Linguistics.

\bibitem[{Aguilar et~al.(2017)Aguilar, Maharjan, Monroy, and
  Solorio}]{aguilar2017multi}
Gustavo Aguilar, Suraj Maharjan, Adrian Pastor~L{\'o}pez Monroy, and Thamar
  Solorio. 2017.
\newblock A multi-task approach for named entity recognition in social media
  data.
\newblock In \emph{Proceedings of the 3rd Workshop on Noisy User-generated
  Text}, pages 148--153.

\bibitem[{Chiu and Nichols(2016)}]{chiu2016named}
Jason~PC Chiu and Eric Nichols. 2016.
\newblock Named entity recognition with bidirectional lstm-cnns.
\newblock \emph{Transactions of the Association for Computational Linguistics},
  4:357--370.

\bibitem[{Collobert et~al.(2011)Collobert, Weston, Bottou, Karlen, Kavukcuoglu,
  and Kuksa}]{collobert2011natural}
Ronan Collobert, Jason Weston, L{\'e}on Bottou, Michael Karlen, Koray
  Kavukcuoglu, and Pavel Kuksa. 2011.
\newblock Natural language processing (almost) from scratch.
\newblock \emph{Journal of Machine Learning Research}, 12(Aug):2493--2537.

\bibitem[{Dyer et~al.(2015)Dyer, Ballesteros, Ling, Matthews, and
  Smith}]{dyer2015transition}
Chris Dyer, Miguel Ballesteros, Wang Ling, Austin Matthews, and Noah~A Smith.
  2015.
\newblock Transition-based dependency parsing with stack long short-term
  memory.
\newblock In \emph{Proceedings of the 53rd Annual Meeting of the Association
  for Computational Linguistics and the 7th International Joint Conference on
  Natural Language Processing (Volume 1: Long Papers)}, volume~1, pages
  334--343.

\bibitem[{Grave et~al.(2018)Grave, Bojanowski, Gupta, Joulin, and
  Mikolov}]{grave2018learning}
Edouard Grave, Piotr Bojanowski, Prakhar Gupta, Armand Joulin, and Tomas
  Mikolov. 2018.
\newblock Learning word vectors for 157 languages.
\newblock In \emph{Proceedings of the International Conference on Language
  Resources and Evaluation (LREC 2018)}.

\bibitem[{Huang et~al.(2015)Huang, Xu, and Yu}]{huang2015bidirectional}
Zhiheng Huang, Wei Xu, and Kai Yu. 2015.
\newblock Bidirectional lstm-crf models for sequence tagging.
\newblock \emph{arXiv preprint arXiv:1508.01991}.

\bibitem[{Lafferty et~al.(2001)Lafferty, McCallum, and
  Pereira}]{lafferty2001conditional}
John~D Lafferty, Andrew McCallum, and Fernando~CN Pereira. 2001.
\newblock Conditional random fields: Probabilistic models for segmenting and
  labeling sequence data.
\newblock In \emph{Proceedings of the Eighteenth International Conference on
  Machine Learning}, pages 282--289. Morgan Kaufmann Publishers Inc.

\bibitem[{Lample et~al.(2016)Lample, Ballesteros, Subramanian, Kawakami, and
  Dyer}]{lample2016neural}
Guillaume Lample, Miguel Ballesteros, Sandeep Subramanian, Kazuya Kawakami, and
  Chris Dyer. 2016.
\newblock Neural architectures for named entity recognition.
\newblock In \emph{Proceedings of NAACL-HLT}, pages 260--270.

\bibitem[{Limsopatham and Collier(2016)}]{limsopatham2016bidirectional}
Nut Limsopatham and Nigel Collier. 2016.
\newblock Bidirectional lstm for named entity recognition in twitter messages.
\newblock In \emph{Proceedings of the 2nd Workshop on Noisy User-generated Text
  (WNUT)}, pages 145--152.

\bibitem[{Ling et~al.(2015)Ling, Dyer, Black, Trancoso, Fermandez, Amir,
  Marujo, and Luis}]{ling2015finding}
Wang Ling, Chris Dyer, Alan~W Black, Isabel Trancoso, Ramon Fermandez, Silvio
  Amir, Luis Marujo, and Tiago Luis. 2015.
\newblock Finding function in form: Compositional character models for open
  vocabulary word representation.
\newblock In \emph{Proceedings of the 2015 Conference on Empirical Methods in
  Natural Language Processing}, pages 1520--1530.

\bibitem[{Mikolov et~al.(2018)Mikolov, Grave, Bojanowski, Puhrsch, and
  Joulin}]{mikolov2018advances}
Tomas Mikolov, Edouard Grave, Piotr Bojanowski, Christian Puhrsch, and Armand
  Joulin. 2018.
\newblock Advances in pre-training distributed word representations.
\newblock In \emph{Proceedings of the International Conference on Language
  Resources and Evaluation (LREC 2018)}.

\bibitem[{Shen et~al.(2017)Shen, Yun, Lipton, Kronrod, and
  Anandkumar}]{shen2017deep}
Yanyao Shen, Hyokun Yun, Zachary Lipton, Yakov Kronrod, and Animashree
  Anandkumar. 2017.
\newblock Deep active learning for named entity recognition.
\newblock In \emph{Proceedings of the 2nd Workshop on Representation Learning
  for NLP}, pages 252--256.

\bibitem[{Srivastava et~al.(2014)Srivastava, Hinton, Krizhevsky, Sutskever, and
  Salakhutdinov}]{srivastava2014dropout}
Nitish Srivastava, Geoffrey Hinton, Alex Krizhevsky, Ilya Sutskever, and Ruslan
  Salakhutdinov. 2014.
\newblock Dropout: A simple way to prevent neural networks from overfitting.
\newblock \emph{The Journal of Machine Learning Research}, 15(1):1929--1958.

\bibitem[{Sundermeyer et~al.(2012)Sundermeyer, Schl{\"u}ter, and
  Ney}]{sundermeyer2012lstm}
Martin Sundermeyer, Ralf Schl{\"u}ter, and Hermann Ney. 2012.
\newblock Lstm neural networks for language modeling.
\newblock In \emph{Thirteenth Annual Conference of the International Speech
  Communication Association}.

\bibitem[{Yang et~al.(2016)Yang, Salakhutdinov, and Cohen}]{cohenmulti}
Zhilin Yang, Ruslan Salakhutdinov, and William Cohen. 2016.
\newblock Multi-task cross-lingual sequence tagging from scratch.
\newblock \emph{arXiv preprint arXiv:1603.06270}.

\end{thebibliography}
\bibliographystyle{acl_natbib}

\end{CJK*}
\end{document}